\def\year{2020}\relax
\documentclass[letterpaper]{article} 
\usepackage{aaai20}  
\usepackage{times}  
\usepackage{helvet} 
\usepackage{courier}  
\usepackage[hyphens]{url}  
\usepackage{graphicx} 
\usepackage{graphicx}  
\usepackage{algorithm2e}
\usepackage{todonotes}
\usepackage{tikz}
\usepackage{standalone}
\usepackage{footmisc}
\usepackage{enumitem}
\usetikzlibrary{decorations.pathreplacing}

\makeatletter
\newcommand{\removelatexerror}{\let\@latex@error\@gobble}
\makeatother
\title{Automated Utterance Generation }
\author{\\ \Large \textbf{Soham Parikh \textsuperscript{\rm 1}\thanks{Work was done during an internship at PassageAI}, Quaizar Vohra \textsuperscript{\rm 2}, Mitul Tiwari \textsuperscript{\rm 2}}\\ 
\textsuperscript{\rm 1}University of Pennsylvania, Philadelphia, PA USA \\
\textsuperscript{\rm 2}Passage AI,  
Mountain View, CA USA\\
\textsuperscript{\rm 1}sohamp@seas.upenn.edu, 
\textsuperscript{\rm 2}\{quaizar,mitul\}@passage.ai 
}

\begin{document}

\maketitle

\begin{abstract}
Conversational AI assistants are becoming popular and question-answering is an important part of any conversational assistant. Using relevant utterances as features in question-answering has shown to improve both the precision and recall for retrieving the right answer by a conversational assistant. Hence, utterance generation has become an important problem with the goal of generating relevant utterances (sentences or phrases) from a knowledge base article that consists of a title and a description. However, generating good utterances usually requires a lot of manual effort, creating the need for an automated utterance generation. In this paper, we propose an utterance generation system which $1$) uses extractive summarization to extract important sentences from the description, $2$) uses multiple paraphrasing techniques to generate a diverse set of paraphrases of the title and summary sentences, and $3$) selects good candidate paraphrases with the help of a novel candidate selection algorithm. 
\end{abstract}


\section{Introduction}

Utterance generation is an important problem in Question-Answering, Information Retrieval, and Conversational AI Assistants. 
Voice assistants like Alexa, Google Assistant, Apple Siri, and Microsoft Cortana are proliferating and now billions of devices have these voice assistants. Any conversational skill developed for these devices needs to understand various ways an end user is asking a question, and be able to respond accurately. While voice assistants are becoming very common, chatbots and conversational interfaces are getting adopted for various conversational automation use cases such as website assistants, customer service automation and IT and enterprise service automation.
Question-answering is an important part of any conversational automation use case. It is critical that a conversational assistant understands various ways that could be used in asking the same question, essentially paraphrases of the later. Using relevant utterances as features in a question-answering system has shown to improve the accuracy both in terms of precision and recall to retrieve the right answer. 



In this paper we address the problem of utterance generation in the context of conversational virtual assistant and question-answering. In case of question-answering, we generate utterances for questions (for example, Frequently Asked Questions or FAQs) so that we can identify the right answer for the corresponding question even if the question is asked in many different ways. We propose an ensemble method for the utterance generation problem with a novel candidate selection algorithm. 
Our method first uses extractive summarization to extract important sentences from the description. Second, we use multiple paraphrase generation techniques to generate a diverse set of paraphrases of the title and summary sentences. Some of the techniques we use are full sentence backtranslation, noun/verb phrase backtranslation using constituency parsing, synonym replacement and phrase replacement for paraphrase generation. We use an ensemble method to combine all these different paraphrasing techniques. Finally, we use a novel candidate selection algorithm that utilizes a combination of filtering and de-duplication techniques to select a good set of utterances by leveraging some of the latest contextual embedding techniques such as BERT~\cite{DBLP:conf_naacl_DevlinCLT19} to find semantically similar utterances and to filter out unrelated utterances and remove duplicate utterances. Our experimental results demonstrate that our ensemble method performs well. The main contributions of this paper are as follows.
\begin{itemize}
\item First, we propose an ensemble method for the utterance generation problem, which combines many approaches to utterance generation, and is scalable to add new approaches.  
\item Second, a novel candidate selection algorithm that uses a combination of filtering and de-duplication techniques to select generated utterances.
\item Third, we adopt recent advances in large scale pre-trained contextual embeddings like BERT and Universal Sentence Encoding for candidate selection to find good utterances. 
\item Finally, we demonstrate through extensive experiments that the proposed techniques work well for the utterance generation problem. 
\end{itemize}

The rest of the paper is organized as follows. 
Section~\ref{relwork} describes related work and gives contexts around our work. 
Section~\ref{probstmt} defines the problem of utterance generation addressed in this paper.
Section~\ref{propmeth} discusses the proposed solution and algorithms for the utterance generation problem.
Section~\ref{experiments} describes the experimental setup and discusses the results.
Finally, we conclude in Section~\ref{conclusion}.

\section{Related Work} \label{relwork}

Paraphrases are sentences/phrases which contain different words or different sequence of words but have the same meaning. However, as explained in \cite{Bhagat2013WhatIA}, loosely equivalent and semantically similar sentences/phrases can also be considered as paraphrases. There have been mainly three lines of work for generating paraphrases. The first line of work uses statistical and rule based methods \cite{DBLP:conf/acl/FaderZE13} to mine paraphrase pairs, typically for short phrases from large monolingual \cite{Barzilay:2001:EPP:1073012.1073020} or bilingual corpora \cite{bannard-callison-burch-2005-paraphrasing}. While these methods extract a large set of paraphrase pairs (\textit{e.g.,} the PPDB Corpus \cite{ganitkevitch-etal-2013-ppdb}), the phrases are mostly short. To overcome this limitation, the second line of work focuses on neural models for paraphrase generation. \cite{mallinson-etal-2017-paraphrasing} use neural machine translation to generate paraphrases by first translating the phrase from English to multiple translations in the reference language (\textit{e.g.,} German) and translate these translations back to English. \cite{prakash-etal-2016-neural} and \cite{DBLP:conf/emnlp/LiJSL18} use sequence-to-sequence networks to directly generate paraphrases of a given input sentence. To encourage diversity among the paraphrases generated, \cite{DBLP:conf/aaai/GuptaASR18} and \cite{kumar-etal-2019-submodular} propose different methods based on sequence to sequence models. The third line of work uses paraphrasing as an intermediate stage for the task of Question Answering \textit{e.g.,} \cite{DBLP:conf/acl/BerantL14}. However, these papers focus on their end goal of Question Answering and they don't evaluate the quality of paraphrases themselves in their experiments.

There are mainly 2 limitations which we observed while experimenting with the methods presented in these papers: either they lacked variety in their paraphrases or they failed to generate relevant paraphrases when used in the domains we are interested in, \textit{e.g.,} customer service automation, possibly due to lack of sufficient domain-specific data.

To generate a rich variety of paraphrases, our paper uses multiple techniques like back-to-back machine translation and replacement using PPDB \cite{ganitkevitch-etal-2013-ppdb} and WordNet \cite{Miller:1995:WLD:219717.219748} resources. We then use a candidate selection algorithm leveraging recent advances in contextual embeddings, ( \cite{DBLP:conf_naacl_DevlinCLT19} and \cite{DBLP:journals/corr/abs-1803-11175}) to select high quality paraphrases with strong semantic similarity with the input sentence. To our knowledge, there is only one other paper which uses a candidate selection algorithm \cite{DBLP:conf/naacl/DuboueC06} where they use a classifier for selecting good paraphrases. Moreover the focus is on Question Answering and hence there is no evaluation of the quality of their paraphrases themselves.

\section{Problem Formulation} \label{probstmt}
A knowledge base article (such as FAQs and manuals) usually consists of a \textit{title} and an associated \textit{description}. A user who needs help with a particular issue can frame the same issue in different ways. For example, a user who wants to pay their bill can use ``How do I pay my bill?", ``I want to pay my bill", ``I wish to settle my dues". Whereas the article can be titled as ``Pay your bill". Information Retrieval based models lack recall when the words chosen by a user are different from the article but are semantically related.  
Enriching articles with utterances that are semantically similar to their content have shown to significantly improve recall and precision of IR based models in our experiments.


This work addresses the problem of automatically generating utterances for a given article, which can be further curated and used by human annotators to prepare a final list of reference utterances for the article. The method described in this work first uses summarization techniques to find important sentences in an article. Next, paraphrase generation techniques are used to generate many candidate utterances for each of these sentences as well as the title. Finally, a novel candidate selection algorithm which performs filtering and de-duplication is used to discard candidates which do not make sense or are not semantically similar to the original sentence and to remove duplicate paraphrases.

\section{Proposed Method}\label{propmeth}

\begin{figure*}[ht]
    \centering
    \includegraphics[width=2.0\columnwidth]{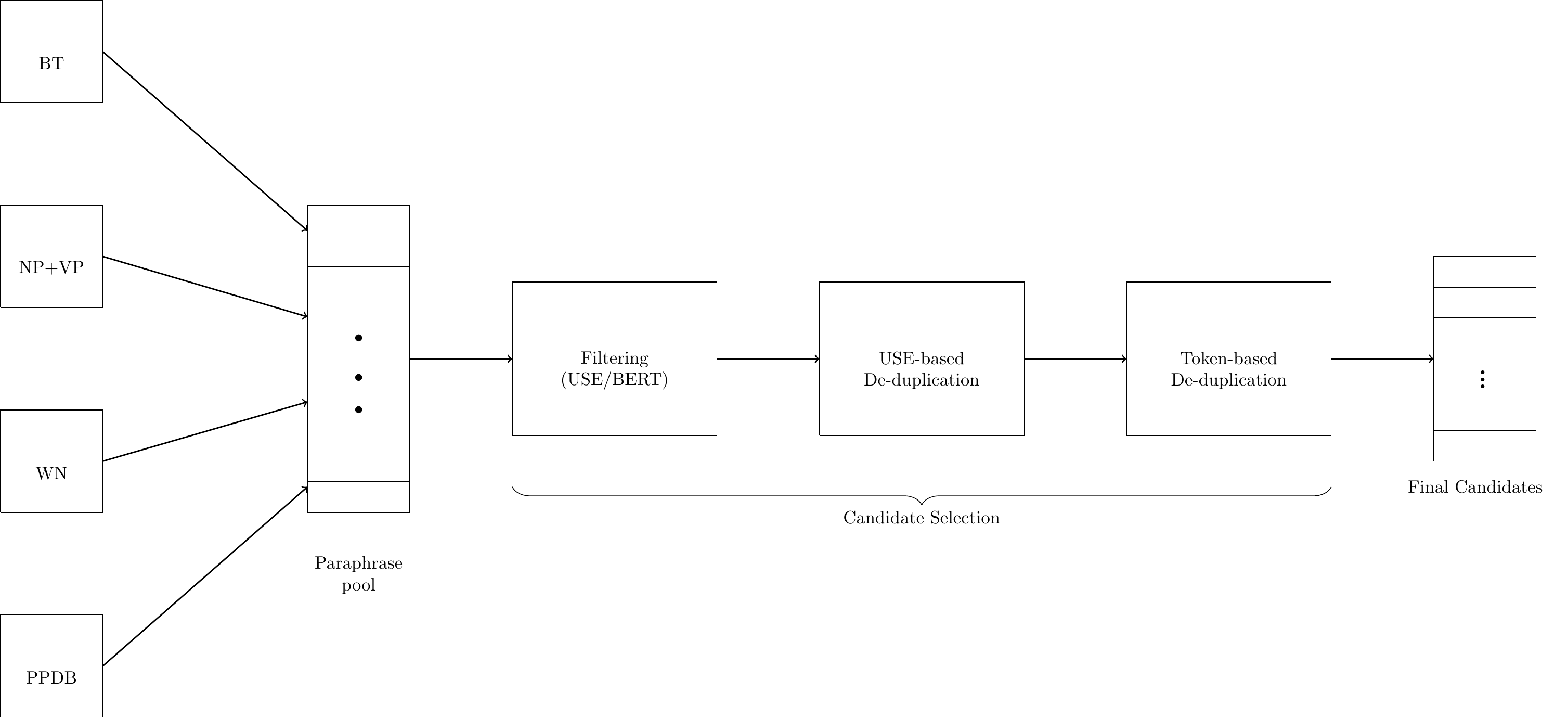}
    \caption{A figure representing the generation and candidate selection pipeline}
\end{figure*}

A conversational assistant user can refer to an article using a question which is either a paraphrase of the title or is related to the text present in the description. The former motivates the need for paraphrase generation. However, descriptions can be long and often contain sentences that users don't refer to. Hence, we use extractive summarization to select important sentences from the description, following which we generate paraphrases for each of the extracted sentences as well as the title of an article. Our aim here is to generate a diverse set of paraphrases for a sentence and hence, we choose to adapt the overgenerate and selection paradigm. We first use multiple paraphrasing techniques to generate a large pool of paraphrases following which we use a novel candidate selection algorithm to select useful and relevant paraphrases for each input sentence. Next, as parts of utterance generation, we will describe our methods for Paraphrase Generation, Candidate Selection and Summarization. 


\subsection{Paraphrase Generation}

We use many different methods for generating paraphrases such as (1) full backtranslation, (2) noun/verb phrase backtranslation using constituency parsing, (3) synonym replacement, and (4)  phrase replacement.
\begin{itemize}
    \item \textbf{Full Backtranslation (BT):} Inspired from \cite{mallinson-etal-2017-paraphrasing}, we use neural machine translation models for generating paraphrases. We first generate multiple German translations of the input English sentence. For each of the German translations, we generate multiple English translations. In order to generate multiple translations, we use beam search at the time of decoding. We also experimented with Czech, however, German seemed to work better for us. 
    \item \textbf{Noun/Verb Phrase Backtranslation (NP/VP):} Backtranslating an entire sentence can often generate lots of duplicate paraphrases, especially when the input sentence is long. Hence, we also generate paraphrases for only a certain meaningful phrase from the input sentence. We use the Berkeley Neural Parser \cite{Kitaev-2018-SelfAttentive} to perform constituency parsing and extract all noun and verb phrases from the input sentence. For each of these extracted phrases, we generate backtranslations and replace the phrase in the original sentence with its respective backtranslations. 
    
    \item \textbf{Synonym Replacement (WN):} Often times, paraphrasing involves replacing a single word with another having equivalent meaning in the context. To account for this, we find synonyms for words in the input sentence from synsets obtained using WordNet \cite{Miller:1995:WLD:219717.219748} and replace the word with its synonym. We do not consider words that are stopwords, whose Part-of-Speech tag belongs to a manually curated list of tags or that are less than 3 characters long. 
    
    \item \textbf{Phrase replacement (PPDB):} WordNet usually contains synonyms for only single words, whereas noun and verb phrase backtranslation generate paraphrases for only certain types of phrases. PPDB \cite{ganitkevitch-etal-2013-ppdb} is a database of paraphrases of commonly occurring phrases, extracted from a bilingual corpus. We use this resource to replace all matching phrases from the input sentence with their paraphrases.
\end{itemize}
\begin{table*}[ht]
    \centering
    \begin{tabular}{c|c|c}
         \textbf{Source} & \textbf{Generated 1} & \textbf{Generated 2}\\
         \hline
         how to resume the preset speed ? & how can i restart the default speed ? & how to recover the preset speed ? \\
         \hline
         how do i activate voice commands? & how do i activate the speech command ? & how do i activate voice control ?\\
         \hline
         change your payment method & payment method amendment & switch your payment method \\
         \hline
         credit limit increases & credit bound increase & raising the credit limit \\
         \hline
         when can i rely on icc? & when can i be dependent upon icc ? & when can i count on icc ?
    \end{tabular}
    \caption{Examples of the some of the useful paraphrases generated by our method. }
    \label{tab:my_label}
\end{table*}
\subsection{Candidate Selection}\label{filterdedup}
Using multiple techniques for paraphrasing generates a large pool of paraphrases which could potentially contain sentences that are semantically different from the input sentence or synonyms replaced in the wrong context as well as duplicates of the title and each other. This necessitates a method to select relevant candidate paraphrases. As part of our candidate selection algorithm, we first remove the irrelevant sentences using a filtering mechanism, following which we use a de-duplication method to remove duplicates. 

\subsubsection{Filtering}
The goal of filtering is to remove paraphrases that are not semantically equivalent. We describe two different filtering methods that we experiment with. 
\begin{itemize}
    \item \textbf{USE-based:} We use the Universal Sentence Encoder \cite{DBLP:journals/corr/abs-1803-11175} to get vector representations of the input sentence and the paraphrase and compute the cosine similarity between them. If the cosine similarity between the representations is less than $0.5$, the paraphrase is considered to be semantically different and is discarded. Analogously, if the similarity is greater than $0.95$, the paraphrase is considered to be a duplicate of the input and hence, also discarded. These thresholds were fine-tuned after experimenting with different thresholds on a set of positive and negative paraphrase pairs.  
    
    \item \textbf{BERT-based:} USE-based similarity determines the semantic similarity between two sentences, however, it does not explicitly tell us if the sentences are semantically equivalent. For similar sentences (\textit{e.g.,} sentences with high word overlap) that are not semantically equivalent, USE-based filtering fails to give the desired result. In order to improve precision of filtering, we use a paraphrase detection model based on BERT \cite{DBLP:conf_naacl_DevlinCLT19}. This model is trained on labeled pairs from Quora Question Pairs (as given by \cite{DBLP:conf/iclr/WangSMHLB19}), MRPC \cite{dolan2005automatically}, STS Benchmark \cite{cer-etal-2017-semeval} and PAWS \cite{DBLP:conf/naacl/ZhangBH19}. 
    
\end{itemize}

\subsubsection{Deduplication}
In order to remove duplicates, we run the following two algorithms sequentially after the filtering step. Algorithm~\ref{algo1} uses similarity based on USE to de-duplicate the pool, that is, at every step, it finds the paraphrase that has the highest cosine similarity with the original sentence and selects it if it does not have a high similarity with any of the paraphrases already selected. Algorithm~\ref{algo2} focuses on diversifying the final set by selecting the paraphrase with the highest number of unique words at every step. We only consider words that are not stopwords, have a character length of more than 2 and whose POS tags do not belong to a manually curated list of POS tags (such as prepositions, conjunction words, and forms of the verbs ``be" and ``have")

\newcommand{\myalgorithm}{
\begingroup
\removelatexerror
\begin{algorithm*}[H]\label{algo1}
    \SetAlgoLined
     output=[]\;
     \textbf{sort}(\textit{pool});
     sentencoding=\textit{USE(}input\textit{)}\;
     \For{paraphrase in pool}{
     vector=\textit{USE(}paraphrase\textit{)}\;
     \For{paraphrase2 in output}{
      \If{\textit{cosine}(\textit{USE(}paraphrase2\textit{)}, vector) $>$ 0.95}{
      \textbf{break}\;
      }
     }
     \If{\textit{cosine}(vector,sentencoding) $<$ 0.95}{\textbf{append} paraphrase to output\;}
     }
    \caption{Deduplication using USE}
\end{algorithm*}
\begin{algorithm*}[H]\label{algo2}
    \SetAlgoLined
     wordset=\{words in input\}\;
     output=[]\;
     \While{len(output) $<$ k}{
        \textbf{select} paraphrase from pool with most number of unique words\;
        \eIf{no such paraphrase exists}{\textbf{break}\;}{
            output.append(paraphrase)\;
            pool.remove(paraphrase)\;
            wordset.append(new words in input)
        }
     }
    \caption{Word based deduplication}
\end{algorithm*}
\endgroup
}
\myalgorithm


\paragraph{Selection during tie-breaking}
While performing de-duplication, many of the paraphrases generated have just one or two keywords that are different and unique from the input sentence. It is important to select the sentences that are more related to the input sentence and which also are more probable as a sentence. Hence, for each paraphrase, we compute two scores, namely, the similarity between the USE encodings of the input sentence and the paraphrase, and a score computed using the cross entropy loss from the BERT model probabilities. We normalize both of these scores across examples and use the average for tie-breaking.


\begin{table*}[t]
    \centering
    \begin{tabular}{|c|c|c|c|c|}
    \hline
         \textbf{Method}& \multicolumn{2}{c|}{\textbf{Avg. Fraction of useful}} & \multicolumn{2}{c|}{\textbf{Avg. number of useful}} \\
         \hline
         & \textbf{USE} & \textbf{BERT} & \textbf{USE} & \textbf{BERT} \\
         \hline
         BT & \textbf{0.6} & \textbf{ 0.608} & 2.31 & 1.16\\
         \hline
         NP + VP & 0.42 & 0.5 & 3.16 & 1.76\\
         \hline
         BT + VP + NP & 0.40 & 0.5 & 3.37 & 2.78\\
         \hline
         WordNet & 0.216 & 0.315 & 2.79 & 2.73\\
         \hline
         PPDB & 0.26 & 0.298 & 2.31 & 2.03\\
         \hline
         BT + VP + NP + WN + PPDB & 0.24 & 0.44 & \textbf{4.27} & \textbf{5.97} \\
         \hline
         CVAE & 0.05 & - & 0.37 & - \\
         \hline
    \end{tabular}
    \caption{Results for manual evaluation. We explain in Discussion why we don't perform filtering with BERT on CVAE.}
    \label{resulttable}
\end{table*}

\begin{table*}[t]
    \centering
    \begin{tabular}{|c|c|c|c|c|c|c|c|c|c|}
         \hline
         \textbf{Method} & \multicolumn{3}{c|}{\textbf{MRPC}} & \multicolumn{3}{c|}{\textbf{STS}} & \multicolumn{3}{c|}{\textbf{Private}} \\
         \hline
         & \textbf{No CS} & \textbf{USE} & \textbf{BERT} & \textbf{No CS} & \textbf{USE} & \textbf{BERT} & \textbf{No CS} & \textbf{USE} & \textbf{BERT}\\
         \hline
         BT & 0.201 & 0.169 & 0.199 & 0.225 & 0.14 & \textbf{0.239} & 0.217 & 0.205 & \textbf{0.234} \\
         \hline
         NP + VP & 0.345 & \textbf{0.325} & 0.325 & 0.172 & 0.173 & 0.173 & 0.155 & 0.172 & 0.172\\
         \hline
         BT + VP + NP & 0.329 & 0.306 & 0.306 & 0.18 & 0.173 & 0.174 & 0.171 & 0.183 & 0.183 \\
         \hline
         WordNet & \textbf{0.393} & 0.27 & \textbf{0.36} & \textbf{0.3} & \textbf{0.26} & 0.173 & \textbf{0.27}& \textbf{0.256} & 0.173 \\
         \hline
         PPDB & 0.27 & 0.236 & 0.262 & 0.246 & 0.213 & 0.169 & 0.249 & 0.24 & 0.171 \\
         \hline
         BT + NP + VP + WN + PPDB & 0.334 & 0.308 & 0.307 & 0.178 & 0.175 & 0.176 & 0.172 & 0.176 & 0.176 \\
         \hline
         CVAE & 0.089 & 0.058 & - & 0.073& 0.0297 & - & 0.194 & 0.139 & - \\
         \hline
    \end{tabular}
    \caption{Result of BLEU scores on different datasets. No CS indicates no candidate selection algorithm.}
    \label{bleuscores}
\end{table*}

\subsection{Summarization}
To select the important sentences from the description, we use extractive summarization. We experimented with the pre-trained models provided by \cite{DBLP:conf/acl/BansalC18} and \cite{DBLP:conf/naacl/NarayanCL18} and chose to go with the former after evaluating on a private test set. For more details regarding the description of the model, we defer to the original paper. We only summarize a description if it is more than 3 sentences long. Otherwise, we pick all sentences as important sentences. 

\section{Experiments}\label{experiments}

In this section, we focus on evaluating the paraphrase generation method. While metrics like BLEU \cite{Papineni:2002:BMA:1073083.1073135}, ROUGE \cite{lin-2004-rouge} and METEOR \cite{Lavie:2007:MAM:1626355.1626389} have been proposed for automatic evaluation, as pointed out in \cite{callison-burch-etal-2006-evaluation} and \cite{Lavie:2007:MAM:1626355.1626389}, these measures are inadequate since they perform n-gram matching, and do not capture diversity. Since the aim is to produce paraphrases that are as diverse as possible, it is hard to come up with all possible reference sentences. Hence, we focus on results of manual evaluation and for completeness, we also report the BLEU score. We also present evaluation results of the CVAE-based model described in \cite{DBLP:conf/aaai/GuptaASR18}. Note that we only evaluate the paraphrase generation method as we believe that is the main contribution of this work. Since the generated utterances will be further curated and referred to by humans, we believe that evaluation on the end task of information retrieval is more nuanced and should be explored as part of future work.  

\subsection{Dataset}
For manual evaluation, we prepare a test set of $100$ sentences from which useful utterances can be generated, manually chosen from articles in auto manuals, telecom company FAQs, and retail FAQs. We report BLEU score on two publicly available datasets, namely, the STS Benchmark ($371$ sentences) and the MRPC corpus ($273$ sentences) as well as a private dataset ($567$ sentences) which contains sentences from auto manuals along with manually generated paraphrases by upworkers. For training the CVAE-based model, we use the Quora Question Pair dataset along with paraphrase pairs from ComQA \cite{DBLP:journals/corr/abs-1809-09528} and our private dataset. 


\subsection{Evaluation Methodology}
For each input sentence, we generate paraphrases and restrict the number of generated paraphrases to a maximum of 20. For backtranslation, we first generate $5$ German translations and further generate $5$ English translations for each of these, resulting in a total of $25$ backtranslations. For the CVAE-based model, we sample 25 random seeds to generate 25 different paraphrases. We use the final pool obtained from aggregating the paraphrases generated using all methods as an input to the candidate selection.


\subsubsection{Manual Evaluation}
Each of the generated paraphrase is marked as either useful ($1$) or not useful ($0$) by the human annotator. We report two metrics for manual evaluation. The first metric is a precision-based metric which computes the fraction of paraphrases that are useful for each sentence and takes an average of this fraction. The second metric that we report is the absolute number of useful paraphrases generated per input sentence averaged across all input sentences. We report this in Table~\ref{resulttable}.

\subsubsection{Automatic Evaluation}
For each of the generated paraphrase, we use the set of reference sentences to compute BLEU score. The score for an input sentence is the average of the BLEU score over all the generated paraphrases for the sentence. We report the average of this score over all input sentences, for the paraphrase sets obtained both before and after filtering. This is reported in Table~\ref{bleuscores}.

\subsection{Discussion} \label{discussion}
We make multiple observations from Tables~\ref{resulttable} and~\ref{bleuscores}. 
\begin{itemize}
    \item Combining multiple techniques generates more number of useful paraphrases than each of the individual methods. 
    \item Using BERT for filtering improves the precision notably. Although it reduces the number of useful paraphrases for individual methods, it works better when all methods are combined. Analysis of the results using USE filtering showed that it gave high score to irrelevant paraphrases generated using WordNet (lot of overlapping words), resulting in a high fraction of them being included in the final set, while useful paraphrases from other methods were given lower score and not included in final set. 
    BERT-based filtering helped in reducing the total number of WordNet-based paraphrases, allowing paraphrases from other methods to also be included in the final set. 
    \item BLEU scores are not consistent with the human evaluation and hence are not very reliable for our purposes.
    \item The CVAE-based model does not perform as well for our purpose as the other methods. We trained multiple models on different combinations of datasets and saw that the model failed to generate good quality paraphrases for previously unseen sentences. It is for this reason that we did not proceed with experimenting BERT-filtering on CVAE.   
\end{itemize}

\section{Conclusion}\label{conclusion}

In this work, we addressed the problem of utterance generation. We use the title of an article along with sentences extracted using summarization of the description as reference sentences. We propose an ensemble method that uses multiple paraphrasing techniques to generate a set of paraphrases from an input sentence. We also propose some innovative ways of paraphrasing using constituency parsing and noun/verb phrase neural backtranslation. Finally, we developed a novel candidate selection algorithm to filter out bad utterances and remove duplicates for diversity using some of the latest contextual embedding techniques such as BERT and Universal Sentence Encoder. Our experimental results show that our overall approach works well. We tried conditional variation autoencoder (CVAE) based techniques but did not get good utterances generated. In the future, we plan to further fine tune and experiment with sequence-to-sequence models like CVAE for generating utterances.



\fontsize{9.0pt}{10.0pt} \selectfont
\bibliography{references}
\bibliographystyle{aaai}

\end{document}